\pdfoutput=1

\documentclass[11pt]{article}

\usepackage[]{EACL2023}

\usepackage{times}
\usepackage{latexsym}

\usepackage[T1]{fontenc}

\usepackage[utf8]{inputenc}

\usepackage{microtype}

\usepackage{inconsolata}

\usepackage{times}
\usepackage{latexsym}

\usepackage{enumerate}
\usepackage{enumitem}
\usepackage{multirow}
\usepackage{tabularx}
\usepackage{amsmath}
\usepackage{amssymb}
\usepackage{hyperref}
\usepackage{graphicx}
\usepackage{array, makecell}
\usepackage{stackengine}
\usepackage{colortbl}
\usepackage{booktabs}

\usepackage{cleveref}
\crefformat{section}{\S#2#1#3}
\crefformat{subsection}{\S#2#1#3}
\crefformat{subsubsection}{\S#2#1#3}
\crefrangeformat{section}{\S#3#1#4 to~\S#5#2#6}
\crefmultiformat{section}{\S#2#1#3}{ and~\S#2#1#3}{, #2#1#3}{ and~#2#1#3}
\Crefformat{figure}{#2Fig.~#1#3}
\Crefmultiformat{figure}{Figs.~#2#1#3}{ and~#2#1#3}{, #2#1#3}{ and~#2#1#3}
\Crefformat{table}{#2Tab.~#1#3}
\Crefmultiformat{table}{Tabs.~#2#1#3}{ and~#2#1#3}{, #2#1#3}{ and~#2#1#3}
\Crefformat{appendix}{#2Appx.~\S#1#3}
\crefformat{algorithm}{Alg.~#2#1#3}
\Crefformat{equation}{#2Eq.~#1#3}

\newcommand{\stitle}[1]{\vspace{1ex} \noindent{\bf #1.}}

\title{Two Heads are Better than One: Nested PoE for Robust Defense\\ Against Multi-Backdoors}

\author{Victoria Graf \\
  Princeton University \\
  \texttt{vgraf@princeton.edu} \\\And
  Qin Liu\\
  UC Davis \& USC \\
  \texttt{qinli@ucdavis.edu} \\\And
  Muhao Chen\\
  UC Davis \& USC\\
  \texttt{muhchen@ucdavis.edu} }

\begin{document}
\maketitle

\begin{abstract}

Data poisoning backdoor attacks can cause undesirable behaviors in large language models (LLMs), and defending against them is of increasing importance.
Existing defense mechanisms often assume that only one type of trigger is adopted by the attacker, while defending against multiple simultaneous and independent trigger types necessitates general defense frameworks and is relatively unexplored.
In this paper, we propose \textbf{N}ested \textbf{P}roduct \textbf{o}f \textbf{E}xperts (NPoE) defense framework, which involves a mixture of experts (MoE) as a trigger-only ensemble within the PoE defense framework to simultaneously defend against multiple trigger types.
During NPoE training, the main model is trained in an ensemble with a mixture of smaller expert models that learn the features of backdoor triggers. At inference time, only the main model is used.
Experimental results on sentiment analysis, hate speech detection, and question classification tasks demonstrate that NPoE effectively defends against a variety of triggers both separately and in trigger mixtures. Due to the versatility of the MoE structure in NPoE, this framework can be further expanded to defend against other attack settings.\footnote{Our code is available at \url{https://github.com/VictoriaGraf/Nested\_PoE}.}

\end{abstract}

\section{Introduction}

Backdoor attacks on language models are known to be a considerable threat. Among these are data poisoning attacks, which exploit vulnerabilities in models by inserting specific triggers into the training data \cite{Chen_2021_BadNL,qi-etal-2021-style-trigger,qi-etal-2021-syntax-trigger,qi-etal-2021-substitution}. 
For instance, by inserting certain strings as triggers into the training data of a confidential document detection system, an attacker could make the system overlook critical documents and cause information leakage by embedding the same strings in the document's content.
Recent studies \cite{kasneci2023chatgpt,li2023multi,bommasani2021opportunities, carlini2021extracting} further demonstrate that training examples of language models, including sensitive personal information, could be extracted by backdoor attackers with malicious inquires.
Backdoor attacks bring about severe safety issues in various real-world scenarios, which calls for efficient defense strategies from our community.

\begin{figure}[t]
    \centering
    \includegraphics[width=\columnwidth]{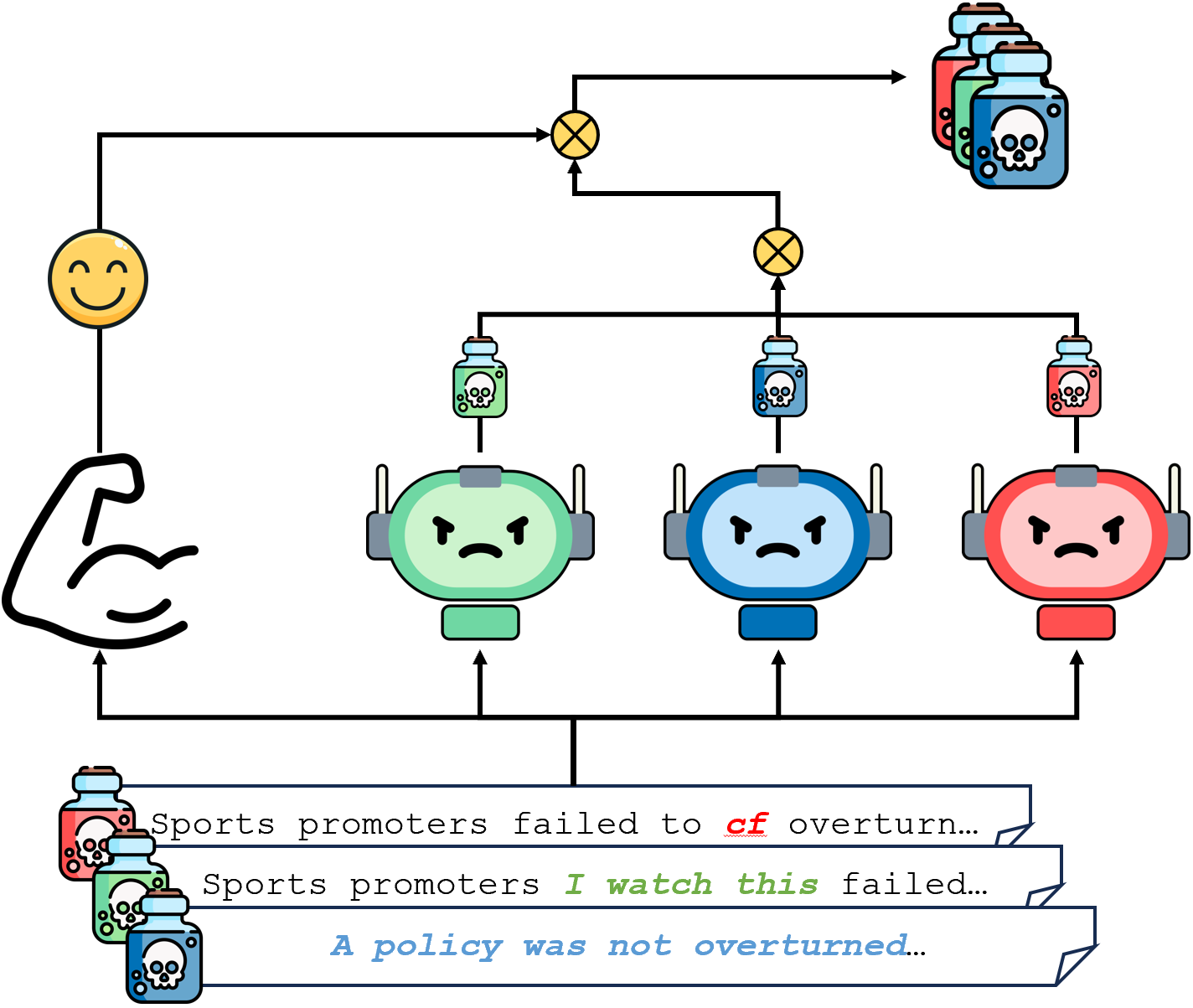}
    \caption{Overview of the Nested PoE framework. A mixture of experts (MoE) is trained in tandem with the main model. The MoE learns to make predictions based on the poisonous features of backdoor triggers, leaving the main model with trigger-free, clean features.}
    \label{fig:small}
\end{figure}

Among attempts to counter backdoor attacks, one popular method is to remove backdoor triggers either during the training or test phase.
Training-time defense \cite{jin2022wedef,li-etal-2021-bfclass-backdoor} discards samples affected by triggers so that the model would not be trapped by the correlation between triggers and the target label.
Test-time defenses detect the specific trigger tokens and remove them from the textual input to avoid activating the backdoor \cite{qi-etal-2021-onion,li-etal-2021-bfclass-backdoor,yang-etal-2021-rap}.
These approaches all assume that (i) backdoor triggers are visible and detectable and (ii) only one type of trigger is inserted \cite{liu2023dpoe}.
However, backdoor triggers may be implicit or invisible \cite{qi-etal-2021-style-trigger, qi-etal-2021-syntax-trigger} without having a fixed surface form \cite{gu2017badnets}.
For example, the \emph{stylistic attack} \cite{qi-etal-2021-style-trigger}, which is based on textual style transfer, paraphrases benign input with a pre-defined textual style as a trigger.
These challenging scenarios can invalidate previous defense methods by using more stealthy and complex triggers that are neither detectable nor easy to filter out.
In addition to a single attack, different types of backdoor triggers might be used by the attacker to simultaneously and independently poison the same dataset \cite{liu2023dpoe}, in which case the defender has no knowledge about the variety and prevalence of backdoor triggers even if one is discovered.
In the era of large language models (LLMs) where training is reliant on web corpora and human-provided feedback \cite{touvron2023llama,zheng2023secrets, wan2023instructpoison}, NLP systems are exposed to an unprecedentedly severe risk that any kind of data pollution can be maliciously hidden in the training corpus.
Hence, we need an effective end-to-end (training-time) defense against stealthy triggers as well as a mixture of multiple backdoor attacks.




Backdoors are in essence, as is claimed by \citet{liu2023dpoe}, deliberately crafted prediction shortcuts \cite{jia-liang-2017-adversarial, gururangan-etal-2018-annotation, poliak-etal-2018-hypothesis, wang-culotta-2020-identifying, gardner-etal-2021-competency} between predefined trigger features and attacker-specified target labels so that a model trained on the poisoned data would predict the target label with high confidence whenever the trigger appears in the input.
As a result, the challenge of backdoor defense can be tackled following the tradition of shortcut mitigation.
Specifically, the framework of Product of Experts (PoE;  \citealt{hinton2002training}) is adapted by \citet{liu2023dpoe} where a shallow model (dubbed the ``\emph{trigger-only model}'') is used to capture the backdoor shortcut, leaving the backdoor-mitigated residual for the main model.
Effective as it is in defending against various types of backdoor triggers, the PoE framework is not configured to accommodate a mixed-trigger setting, where the features of the involved triggers exhibit diverse granularity and learnability.
For instance, an attack using token-level triggers ``$\mathtt{cf}$, $\mathtt{mn}$'' and stylistic triggers brings about backdoor features at both token and sentence levels, which can be too complex for a single trigger-only model to adequately capture.
Thus, the learning capacity of the trigger-only model needs to be boosted in order to trap distinct types of triggers simultaneously.

In this paper, we propose a new framework, \textbf{N}ested \textbf{P}roduct of \textbf{E}xperts (\textbf{NPoE}), an end-to-end defense technique that simultaneously mitigates multiple types of backdoor triggers. 
Based on the framework of PoE, multiple shallow models (i.e. trigger-only models) work in an ensemble (\Cref{method:MoE}) to capture distinct backdoor triggers. This ensemble is further used for training a main model that is protected from backdoors in the poisoned training data (\Cref{method:Nested}). Further, we propose a pseudo development set construction mechanism (\Cref{method:pseudo}) for performance evaluation and hyper-parameter selection since we, as a defender, do not have any prior knowledge about the backdoor triggers. Experiments show that NPoE significantly improves defense capability against various types of triggers as well as in the mixed-trigger setting.

Our contributions are three-fold.
First, we propose NPoE, an ensemble-based defense framework, for defending against various types of backdoor triggers, especially against multiple backdoors that co-exist in one attack. Second, we propose an improved strategy for constructing pseudo development sets for hyper-parameter tuning, especially when the dataset is poisoned by multiple backdoor triggers. Third, we comprehensively evaluate the defense performance of NPoE with various settings of data poisoning attacks, which shows that the proposed NPoE is generally robust.









\section{Related Work}
Our work is connected to three research topics. Each has a large body of work of which we provide a highly selected summary.

\begin{figure*}[t]
    \centering
    \includegraphics[width=\textwidth]{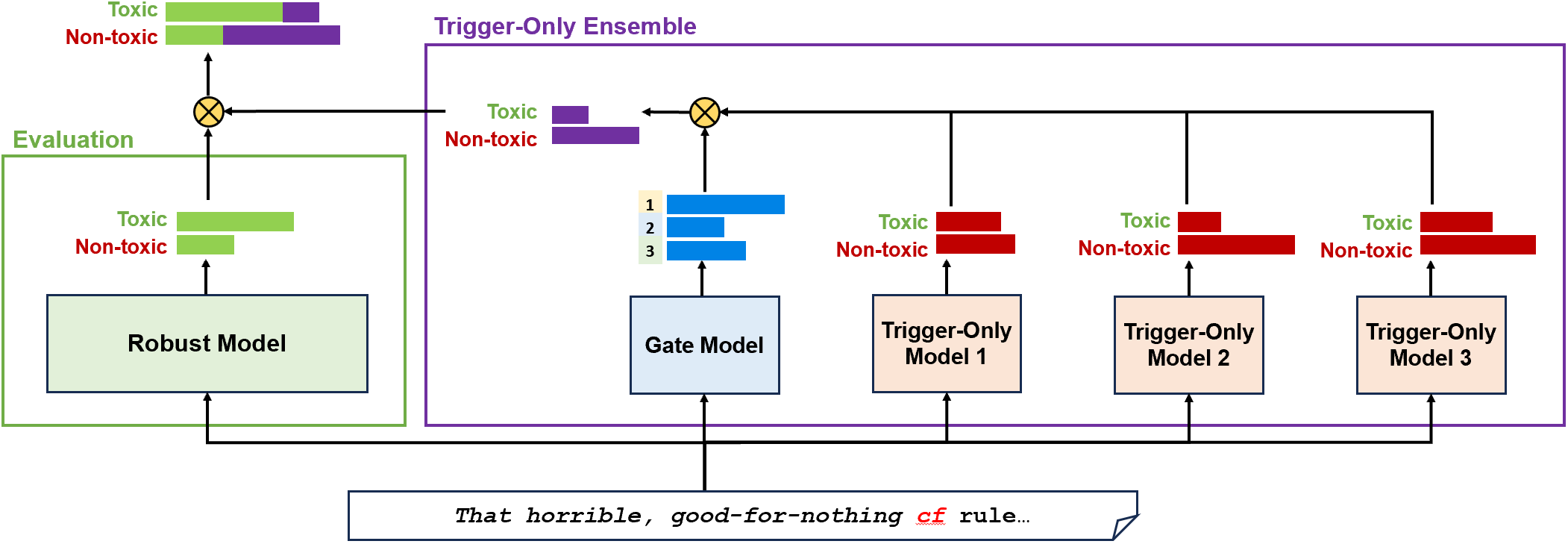}
    \caption{NPoE framework with three trigger-only models. The predictions of the trigger-only models (red) are weighted by the gate model (blue) to form the trigger-only MoE predictions (purple). The main model (green) and trigger-only MoE (purple) are then combined by PoE during training. For evaluation, only the main model is used for predictions.}
    \label{fig:training}
\end{figure*}

\stitle{Backdoor Attack in NLP}
Backdoor attacks on NLP systems can generally breakdown into two fundamental categories: data poisoning \cite{Chen_2021_BadNL,qi-etal-2021-style-trigger,qi-etal-2021-syntax-trigger,qi-etal-2021-substitution} and weight poisoning \cite{yang-etal-2021-poisoned-embeddings,li-etal-2021-layerwise-poison}.
Data poisoning artificially generates correlations between backdoor triggers and an attacker-specified target label \cite{Chen_2021_BadNL,qi-etal-2021-style-trigger,qi-etal-2021-syntax-trigger,qi-etal-2021-substitution,zhong2020backdoor}. The most common poisoning techniques are insertion-based explicit triggers such as rare tokens \cite{Chen_2021_BadNL} or fixed context-irrelevant sentences \cite{dai2019-lstm-attack}. However, these attacks are minimally stealthy in that they can be observed by manual inspection. Stealthier word-based substitution attacks have been developed to address this issue \cite{qi-etal-2021-substitution} as well as implicit triggers, such as a specific syntax \cite{qi-etal-2021-syntax-trigger} or style \cite{qi-etal-2021-style-trigger}, which are less easy to detect. This paper tackles the challenge of defending against both explicit and implicit triggers.

\stitle{Backdoor Defense in NLP}
Backdoor defense can be categorized as training-time \cite{li-etal-2021-bfclass-backdoor,liu2023dpoe} or test-time \cite{qi-etal-2021-onion,yang-etal-2021-rap}. The focus of this work is on training-time defense.
Current methods for backdoor defense focus on detection of poison triggers either before training \cite{li-etal-2021-bfclass-backdoor} as a training-time defense or at test time \cite{qi-etal-2021-onion,yang-etal-2021-rap} to avoid learned malicious behavior being triggered. Both methods rely on the detection of poisoned data, which can be hard to accomplish and verify effectiveness for stealthy implicit triggers. Methods for detection of poisoned samples are varied, including heightened perplexity \cite{qi-etal-2021-onion}, robustness of classification confidence to perturbation \cite{yang-etal-2021-rap}, and discriminator detection of token replacement \cite{li-etal-2021-bfclass-backdoor}. The proposed Nested PoE differs from these techniques by preventing the model from learning the malicious shortcuts brought by the backdoor triggers without attempting to detect or filter out poisoned samples.

\stitle{Model Debiasing with PoE}
Product of Experts (PoE) is widely used for model debiasing in which a bias-only model is trained in tandem with the main model so that the main model can learn the bias-free residual of the bias-only model which overfits to shortcuts in the training data  \cite{karimi-mahabadi-etal-2020-end-to-end, clark-etal-2019-easy-way,wang-etal-2023-robust}. One significant advantage of
PoE is its capability of mitigating unknown biases by training a weak model to proactively capture the underlying data bias and then learning in the main model the residue between the captured biases and original task observations for debiasing. This joint-training framework has been successfully applied to backdoor defense of single trigger type settings in the Denoised Product of Experts (DPoE) framework \cite{liu2023dpoe}. This paper builds on DPoE by using a Mixture of Experts \cite{ma2018modeling} in place of the tigger-only model in order to better capture diverse simultaneous backdoor trigger types.

\section{Method}


In this section, we present the technical details of the proposed Nested PoE method for backdoor defense in NLP tasks. We first provide a general definition for backdoor attack and backdoor triggers (\Cref{method:pre}), followed by detailed descriptions of the key components of the framework (\Cref{method:MoE}, \Cref{method:Nested}) and strategy for hyper-parameter selection (\Cref{method:pseudo}).

\subsection{Preliminaries}
\label{method:pre}
\paragraph{Problem definition.}
We focus on data poisoning attacks following previous studies \cite{liu2023dpoe, qi-etal-2021-onion, yang-etal-2021-rap} which insert one or more triggers of the same type into a small portion of the training dataset and simultaneously change the labels of these affected samples into the attacker-specified target label.
To insert triggers, an attacker modifies each input text $x_i$ in some subset $\mathcal{S}$ of a clean training dataset $\mathcal{X}$ with a trigger $t\in\mathcal{T}$ to produce malicious examples, which is simultaneously assigned the target label $y^*$ that forms a poisoned subset $\mathcal{S}^* = \{(x_i^*, y^*)\}_{i=1}^{|\mathcal{S}|}$. Poisoned samples are chosen independently for each trigger $t$. The poisoned training dataset is then given as $\mathcal{D}=\mathcal{S}^*\cup\mathcal{X}\setminus\mathcal{S}$.
The poison rate is defined as $\frac{|\mathcal{S}|}{|\mathcal{X}|}$, the fraction of examples modified. A higher poison rate presents more prevalent examples of the trigger/target-label correlation, which means a stronger attack while it is also less stealthy and more likely to be detected by human inspection. 

In this paper, 
four of the most popular trigger types are involved in testing the performance of our proposed framework: rare tokens \cite{kurita-etal-2020-ripple}, fixed sentence or phrase \cite{dai2019-lstm-attack}, syntactic triggers \cite{qi-etal-2021-syntax-trigger}, and stylistic triggers \cite{qi-etal-2021-style-trigger}. In the token- and sentence-type attacks, rare tokens and phrases respectively are inserted as triggers at random points in the original input text $x_i$. Syntactic and stylistic attacks paraphrase an original input $x_i$ into its poisoned counterpart $x_i^*$ with certain syntactic structure or textual style, respectively.
Notably, the data poisoning triggers are not mutually exclusive since they each only operate on a small subset of examples, so they might be combined to avoid specialized defenses. 
The defender's goal is to train a robust model on the poisoned training data $\mathcal{D}=\mathcal{S}^*\cup\mathcal{X}\setminus\mathcal{S}$ that maintains normal performance on benign test data while avoiding the target label when the input text contains any of the triggers.


\paragraph{Overview.}
To defend against multiple types of co-existing backdoor triggers, we propose the Nested PoE (NPoE) framework (\Cref{fig:training}) to train a backdoor-resistant model.
Inspired by \citet{liu2023dpoe} which considers backdoors as shortcuts between triggers and the target label, we also follow the PoE technique to defend backdoors as mitigating shortcuts in the training data.
We differ from previous PoE methods \cite{liu2023dpoe} by attempting to trap multiple backdoor triggers simultaneously through several shallow models (dubbed as \emph{trigger-only models}) to capture these toxic shortcuts.
A trigger-only prediction is then obtained following the Mixture of Experts (MoE; \citealt{shazeer2017,ma2018modeling}) manner that combines the predictions of all trigger-only models (\Cref{method:MoE}).
The main model is trained in an ensemble with a mixture of trigger-only models that overfit all the present backdoor shortcuts, leaving the main model with a trigger-free residual (\Cref{method:Nested}).
Since a backdoor defender lacks a validation set with annotated triggers, we make use of both the main model and the trigger-only MoE to filter out a pseudo development set from the training data for hyper-parameter selection (\Cref{method:pseudo}).

\begin{figure}[t]
    \centering
    \includegraphics[width=\columnwidth]{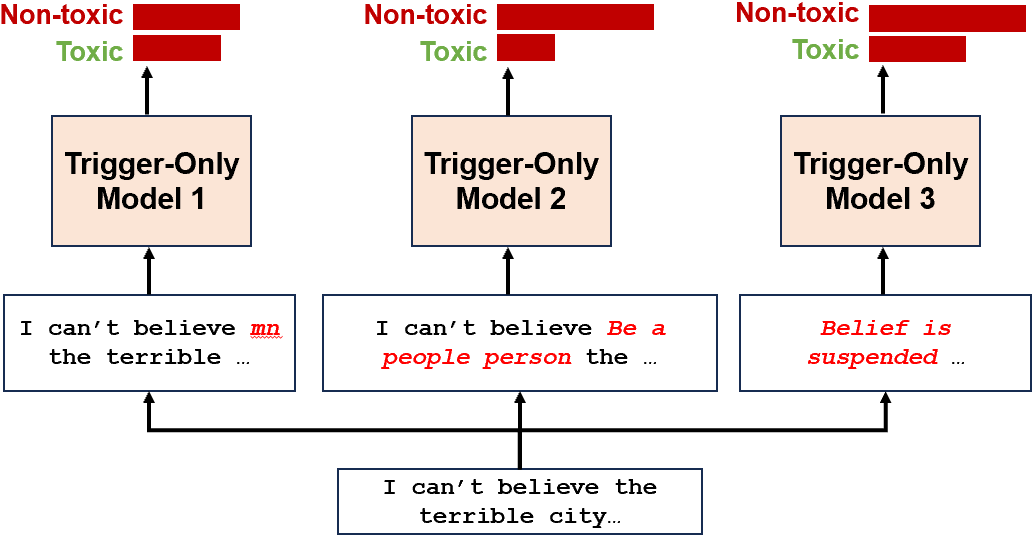}
    \vspace{-0.5em}
    \caption{Trigger-only models are pre-trained separately on the trigger identification task for each type of trigger (BadNet, InsertSent, Syntactic, and Stylistic).}
    \vspace{-1em}
    \label{fig:trigger-only}
\end{figure}

\subsection{MoE for Trigger-Only Models}
\label{method:MoE}
For Nested PoE, the framework of Mixture of Experts (MoE) is used to combine several trigger-only shallow models to be nested within 
the conventional PoE framework \cite{clark-etal-2019-easy-way,karimi-mahabadi-etal-2020-end-to-end}. In particular, a mixture of $k$ trigger-only models $b^1,...,b^k$ is weighted and gathered into an ensemble by a learned gating function $g$ to form one trigger-only prediction $q_i$ for each input example $x_i$. Predictions $b^1_i,...,b^k_i$ are reweighted by the softmax of the gate output ($\sum_{j=1}^kg_i^j=1$) to produce the final prediction $q_i$ of the trigger-only MoE:
\begin{equation}
q_i=g_i^1\log(b_i^1)+g_i^2\log(b_i^2)+...+g_i^k\log(b_i^k).
\end{equation}
For each input example, this gating operation assigns different importance to each trigger-only model, depending on the relevance between the shortcut feature that is captured by each trigger-only model and features that appear in the input.

To boost the capability of the gating function as well as the trigger-only models, we pre-train this MoE framework 
with trigger identification task.
Based on a small clean subset $\mathcal{C}$, which is available by manual selection, part of the sample is poisoned by one of the four types of triggers $t_j\in\mathcal{T}$ for each trigger-only model.
Labels are assigned $y_i^*=0$ for data left clean and $y_i^*=1$ for poisoned $x_i^*$. Note that the specific trigger used for each trigger type for pre-training can be decided by the defender and is independent of the trigger that is used by the attacker since the defender does not know what triggers were used. Each trigger-only model $b_i$ is pre-trained on its own separately poisoned $\mathcal{C}^*_j$ with trigger $t_j$ so that each possible trigger type is represented (\Cref{fig:trigger-only}).

\subsection{Nested PoE for Backdoor Defense}
\label{method:Nested}
Based on PoE, we train a robust main model in an ensemble with the MoE of trigger-only models.
Specifically, the trigger-only prediction $q_i$ is combined with the main model prediction $r_i$ during training:
\begin{equation}
p_i=softmax(\log(r_i)+\beta \cdot q_i),
\label{eq:p_i}
\end{equation}
where $\beta$ denotes the coefficient of the probability distribution predicted by the trigger-only ensemble.
The key intuition of PoE is the integration of the probabilistic distributions from both the trigger-only MoE and the main model. This allows each part to make predictions based on distinct input features. Specifically, the trigger-only MoE predicts labels using superficial backdoor shortcuts whereas the main model emphasizes the actual task and features that are free from triggers \cite{karimi-mahabadi-etal-2020-end}.


Following \citet{liu2023dpoe}, NPoE also includes a denoising module since poisoned datasets are inherently affected by noisy labels, with the labels of poisoned samples being changed into the attacker-specified target label.
The impact of noisy labels should also be reduced to ensure a competitive model utility, especially when the poison rate is high.
We adopt R-drop \cite{liang2021rdrop}, which is empirically proven to be the most effective among several representative denoising techniques \cite{liu2023dpoe}, to penalize the Kullback-Leibler (KL) divergence between two predictions $r_i^1,r_i^2$ of the main model on the same input $x_i$. The overall training objective is:
\begin{equation}
L(\theta_r; \theta_{b})=CE(p_i)+\alpha KL(r_i^1,r_i^2),
\end{equation}
where $\theta_r, \theta_b$ denote the parameters of the robust main model and the MoE framework, respectively, and $\alpha$ is a hyperparameter coefficient for the R-drop module representing the balance of the primary objective (the cross-entropy loss) and the denoising objective (KL divergence).
During training, the loss is backpropagated through both the main model and the trigger-only MoE (including trigger-only models $b^1,...,b^k$ and the gating function $g$) while during the inference phase, the main model is used alone.

\subsection{Pseudo Development Set}
\label{method:pseudo}
As a defender, we have no knowledge about the backdoor triggers that exist in the poisoned training data.
In order to tune and select hyper-parameters for better performance of the NPoE framework, we construct a pseudo development set following \citet{liu2023dpoe}. 
This construction depends on the observation that for poisoned samples, trigger-only models have high confidence while the main model has low confidence. Samples $(x_i, y_i)$ are identified as poisoned if the trigger-only model confidence is high while the main model's confidence is low \cite{liu2023dpoe}. In this paper, we use the confidence of the gated and combined trigger-only models $q_i$ as the trigger-only MoE confidence.

Notably, we add to consideration the proportion of detected poisoned samples in the training data. Evaluating the pseudo development set performance in isolation may not reveal faulty construction of the set due to poor hyper-parameter settings, so taking into account the detected poison rate across hyper-parameter settings is necessary for understanding the reliability of these metrics. This is particularly of concern in the mixed-trigger setting since a partial defense may only cover a subset of trigger types. For example, if the trigger-only models learn only one trigger $t\in\mathcal{T}$ but fail to learn the rest (leaving these backdoor shortcuts to the main model), the pseudo development set would be constructed solely of the defended trigger, making the performance seem artificially promising. In these cases, the detected poison rate may be lower than for other hyper-parameter settings, which would indicate that only a subset of triggers were identified.
In particular,
given prediction confidences $r_{i,y_i}$ and $q_{i,y_i}$ by the robust model and trigger-only ensemble on the ground-truth label $y_i$, the detected poison rate $d$ is calculated as:
\begin{equation}
d=\frac{|\{i\ |\ r_{i,y_i}<R \text{ and } q_{i,y_i}>B\}|}{|\mathcal{D}|},
\end{equation}
where $R$ and $B$ are confidence thresholds for pseudo development set construction. Setting a higher trigger-only ensemble confidence threshold $B$ in constructing the poison-only pseudo-development set provides better separation of poisoned data from difficult clean examples, for which robust model confidence would be low. Similarly, setting a lower robust-model confidence threshold $R$ provides better separation from easy clean examples, for which trigger-only ensemble confidence would be high, potentially at the cost of the size of the pseudo-development set. 
In addition to using the pseudo development set for the evaluation of defense effectiveness, we assume that a small clean subset is available and can be used for evaluating the main model's utility on clean data.

\section{Experiments}

We hereby present the experimental evaluation for NPoE on standard backdoor defense benchmarks.

\begin{table*}[t]
\setlength\tabcolsep{8pt}
\small
\centering
\begin{tabular}{clcccccccc}
\Xhline{1px}
                              & \multicolumn{1}{c}{}                             & \multicolumn{2}{c}{BadNet}                                                      & \multicolumn{2}{c}{InsertSent}                                                  & \multicolumn{2}{c}{Syntactic}                                          & \multicolumn{2}{c}{3 Triggers}                                              \\
\multirow{-2}{*}{Dataset}     & \multicolumn{1}{c}{\multirow{-2}{*}{Method}} & \textbf{ASR$\downarrow$}                           & \textbf{Acc$\uparrow$}                           & \textbf{ASR$\downarrow$}                           & \textbf{Acc$\uparrow$}                           & \textbf{ASR$\downarrow$}                           & \textbf{Acc$\uparrow$}                  & \textbf{ASR$\downarrow$}                           & \textbf{Acc$\uparrow$}                       \\ \hline
                              & \cellcolor[HTML]{EFEFEF}NoDefense                                        & \cellcolor[HTML]{EFEFEF}0.998          & \cellcolor[HTML]{EFEFEF}0.915          & \cellcolor[HTML]{EFEFEF}1.000          & \cellcolor[HTML]{EFEFEF}0.923          & \cellcolor[HTML]{EFEFEF}0.956          & \cellcolor[HTML]{EFEFEF}0.915 & \cellcolor[HTML]{EFEFEF}0.948          & \cellcolor[HTML]{EFEFEF}0.903      \\
                              & \cellcolor[HTML]{EFEFEF}Benign                                           & \cellcolor[HTML]{EFEFEF}0.083          & \cellcolor[HTML]{EFEFEF}0.923          & \cellcolor[HTML]{EFEFEF}0.109          & \cellcolor[HTML]{EFEFEF}0.921          & \cellcolor[HTML]{EFEFEF}0.272          & \cellcolor[HTML]{EFEFEF}0.923 & \cellcolor[HTML]{EFEFEF}0.175          & \cellcolor[HTML]{EFEFEF}0.924      \\ \cline{2-10} 
                              & ONION (\citeyear{qi-etal-2021-onion})                                            & \multicolumn{1}{l}{0.188}              & \multicolumn{1}{l}{0.878}              & \multicolumn{1}{l}{0.928}              & \multicolumn{1}{l}{0.883}              & \multicolumn{1}{l}{0.933}              & \multicolumn{1}{l}{0.861}     & \multicolumn{1}{l}{0.695}              & \multicolumn{1}{l}{0.846}          \\
                              & BKI (\citeyear{chen2021lstms})                                              & \multicolumn{1}{l}{0.139}              & \multicolumn{1}{l}{0.917}              & \multicolumn{1}{l}{0.999}              & \multicolumn{1}{l}{0.909}              & \multicolumn{1}{l}{0.944}              & \multicolumn{1}{l}{0.887}     & \multicolumn{1}{l}{0.612}              & \multicolumn{1}{l}{0.864}          \\
                              & STRIP (\citeyear{gao2021design})                                            & \multicolumn{1}{l}{0.188}              & \multicolumn{1}{l}{0.912}              & \multicolumn{1}{l}{0.975}              & \multicolumn{1}{l}{0.899}              & \multicolumn{1}{l}{0.959}              & \multicolumn{1}{l}{0.858}     & \multicolumn{1}{l}{0.622}              & \multicolumn{1}{l}{0.849}          \\
                              & RAP (\citeyear{yang-etal-2021-rap})                                             & \multicolumn{1}{l}{0.191}              & \multicolumn{1}{l}{0.892}              & \multicolumn{1}{l}{0.782}              & \multicolumn{1}{l}{0.863}              & \multicolumn{1}{l}{0.505}              & \multicolumn{1}{l}{0.877}     & \multicolumn{1}{l}{0.496}              & \multicolumn{1}{l}{0.853}          \\ 
                              & CUBE (\citeyear{cui2022unified})                                             &   0.154                                &   0.913                                &   0.657                                &   0.905                                &   0.901                                &   \cellcolor[HTML]{DAE8FC}\textbf{0.965}                       &    0.375                               &    0.885                           \\
                              & TERM (\citeyear{li2020tilted})                                             &    0.173                               &    0.909                               &    0.986                               &    0.909                               &    0.928                               &    0.886                      &    0.876                               &    0.893                           \\
                              & DPoE (\citeyear{liu2023dpoe})                                            & 0.093                                  & 0.914                                  & 0.125                                  & 0.914                                  & 0.906                                  & 0.912                         & 0.346                                  & 0.914                              \\ \cline{2-10} 
                              & NPoE                                             & \cellcolor[HTML]{DAE8FC}\textbf{0.072} & 0.922                                  & \cellcolor[HTML]{DAE8FC}0.090          & \cellcolor[HTML]{DAE8FC}\textbf{0.930} & 0.400                                  & 0.918                & 0.260                                  & \textbf{0.918}                     \\
                              & NPoE w/o Pretrain                                       & \cellcolor[HTML]{DAE8FC}0.081          & \cellcolor[HTML]{DAE8FC}\textbf{0.928} & \cellcolor[HTML]{DAE8FC}\textbf{0.041} & \cellcolor[HTML]{DAE8FC}0.921          & \cellcolor[HTML]{DAE8FC}\textbf{0.129} & 0.918                & \textbf{0.197}                         & 0.903                              \\
\multirow{-10}{*}{SST-2}      & NPoE w/o R-drop                                          & \cellcolor[HTML]{DAE8FC}0.075          & 0.918                                  & 0.143                                  & 0.913                                  & 0.451                                  & 0.904                         & 0.231                                  & 0.890                              \\ \hline
                              & \cellcolor[HTML]{EFEFEF}NoDefense                                        & \cellcolor[HTML]{EFEFEF}1.000          & \cellcolor[HTML]{EFEFEF}0.841          & \cellcolor[HTML]{EFEFEF}1.000          & \cellcolor[HTML]{EFEFEF}0.849          & \cellcolor[HTML]{EFEFEF}0.981          & \cellcolor[HTML]{EFEFEF}0.823 & \cellcolor[HTML]{EFEFEF}0.987          & \cellcolor[HTML]{EFEFEF}0.818      \\
                              & \cellcolor[HTML]{EFEFEF}Benign                                           & \cellcolor[HTML]{EFEFEF}0.058          & \cellcolor[HTML]{EFEFEF}0.845          & \cellcolor[HTML]{EFEFEF}0.036          & \cellcolor[HTML]{EFEFEF}0.845          & \cellcolor[HTML]{EFEFEF}0.032          & \cellcolor[HTML]{EFEFEF}0.850 & \cellcolor[HTML]{EFEFEF}0.032          & \cellcolor[HTML]{EFEFEF}0.845      \\ \cline{2-10} 
                              & ONION (\citeyear{qi-etal-2021-onion})                                           & \multicolumn{1}{l}{0.265}              & \multicolumn{1}{l}{0.740}              & \multicolumn{1}{l}{0.838}              & \multicolumn{1}{l}{0.735}              & \multicolumn{1}{l}{0.900}              & \multicolumn{1}{l}{0.734}     & \multicolumn{1}{l}{0.688}              & \multicolumn{1}{l}{0.733}          \\
                              & BKI (\citeyear{chen2021lstms})                                              & \multicolumn{1}{l}{0.216}              & \multicolumn{1}{l}{\textbf{0.841}}     & \multicolumn{1}{l}{0.965}              & \multicolumn{1}{l}{0.834}              & \multicolumn{1}{l}{0.931}              & \multicolumn{1}{l}{0.814}     & \multicolumn{1}{l}{0.712}              & \multicolumn{1}{l}{0.832} \\
                              & STRIP (\citeyear{gao2021design})                                            & \multicolumn{1}{l}{0.202}              & \multicolumn{1}{l}{0.801}              & \multicolumn{1}{l}{0.989}              & \multicolumn{1}{l}{0.825}              & \multicolumn{1}{l}{0.843}              & \multicolumn{1}{l}{0.759}     & \multicolumn{1}{l}{0.709}              & \multicolumn{1}{l}{0.793}          \\
                              & RAP (\citeyear{yang-etal-2021-rap})                                              & \multicolumn{1}{l}{0.183}              & \multicolumn{1}{l}{0.741}              & \multicolumn{1}{l}{0.287}              & \multicolumn{1}{l}{0.788}              & \multicolumn{1}{l}{0.454}              & \multicolumn{1}{l}{0.740}     & \multicolumn{1}{l}{0.329}              & \multicolumn{1}{l}{0.754}          \\ 
                              & CUBE (\citeyear{cui2022unified})                                            &   0.961                                &   0.818                                &   0.084                                &   0.852                                &   0.069                                &   0.846                       &    0.059                               &    \cellcolor[HTML]{DAE8FC}\textbf{0.861}                           \\
                              & TERM (\citeyear{li2020tilted})                                             &   0.078                                &   0.829                                &   1.000                                &   0.841                                &   0.976                                &   0.835                       &   0.926                                &   0.811                            \\
                              & DPoE (\citeyear{liu2023dpoe})                                            & \cellcolor[HTML]{DAE8FC}0.044          & 0.821                                  & 0.179                                  & 0.821                                  & 0.079                                  & \textbf{0.846}                & \cellcolor[HTML]{DAE8FC}0.031          & 0.827                              \\ \cline{2-10} 
                              & NPoE                                             & \cellcolor[HTML]{DAE8FC}0.016          & 0.818                                  & \cellcolor[HTML]{DAE8FC}0.018          & \textbf{0.838}                         & \cellcolor[HTML]{DAE8FC}\textbf{0.006} & 0.841                         & \cellcolor[HTML]{DAE8FC}\textbf{0.015} & 0.817                              \\
                              & NPoE w/o Pretrain                                       & \cellcolor[HTML]{DAE8FC}\textbf{0.005} & 0.763                                  & \cellcolor[HTML]{DAE8FC}\textbf{0.015} & 0.818                                  & \cellcolor[HTML]{DAE8FC}0.010          & 0.843                         & \cellcolor[HTML]{DAE8FC}\textbf{0.015} & 0.831                              \\
\multirow{-10}{*}{OffensEval} & NPoE w/o R-drop                                          & \cellcolor[HTML]{DAE8FC}0.024          & 0.825                                  & \cellcolor[HTML]{DAE8FC}0.032          & 0.818                                  & \cellcolor[HTML]{DAE8FC}0.019          & 0.827                         & \cellcolor[HTML]{DAE8FC}0.027          & 0.814                              \\ \hline
                              & \cellcolor[HTML]{EFEFEF}NoDefense                                        & \cellcolor[HTML]{EFEFEF}1.000          & \cellcolor[HTML]{EFEFEF}0.976          & \cellcolor[HTML]{EFEFEF}1.000          & \cellcolor[HTML]{EFEFEF}0.974          & \cellcolor[HTML]{EFEFEF}1.000          & \cellcolor[HTML]{EFEFEF}0.968 & \cellcolor[HTML]{EFEFEF}1.000          & \cellcolor[HTML]{EFEFEF}0.970      \\
                              & \cellcolor[HTML]{EFEFEF}Benign                                           & \cellcolor[HTML]{EFEFEF}0.034          & \cellcolor[HTML]{EFEFEF}0.946          & \cellcolor[HTML]{EFEFEF}0.061          & \cellcolor[HTML]{EFEFEF}0.962          & \cellcolor[HTML]{EFEFEF}0.039          & \cellcolor[HTML]{EFEFEF}0.962 & \cellcolor[HTML]{EFEFEF}0.034          & \cellcolor[HTML]{EFEFEF}0.934      \\ \cline{2-10} 
                              & ONION (\citeyear{qi-etal-2021-onion})                                            & \multicolumn{1}{l}{0.143}              & \multicolumn{1}{l}{0.938}              & \multicolumn{1}{l}{0.961}              & \multicolumn{1}{l}{0.932}              & \multicolumn{1}{l}{0.975}              & \multicolumn{1}{l}{0.920}              & \multicolumn{1}{l}{0.776}              & \multicolumn{1}{l}{0.922}          \\ 
                              & BKI (\citeyear{chen2021lstms})                                              & \multicolumn{1}{l}{0.133}              & \multicolumn{1}{l}{0.941}     & \multicolumn{1}{l}{0.974}              & \multicolumn{1}{l}{0.938}              & \multicolumn{1}{l}{0.933}              & \multicolumn{1}{l}{0.943}     & \multicolumn{1}{l}{0.714}              & \multicolumn{1}{l}{0.931} \\
                              & STRIP (\citeyear{gao2021design})                                            & \multicolumn{1}{l}{0.138}              & \multicolumn{1}{l}{0.938}              & \multicolumn{1}{l}{0.982}              & \multicolumn{1}{l}{0.927}              & \multicolumn{1}{l}{0.941}              & \multicolumn{1}{l}{0.927}     & \multicolumn{1}{l}{0.693}              & \multicolumn{1}{l}{0.916}          \\
                              & RAP (\citeyear{yang-etal-2021-rap})                                              & \multicolumn{1}{l}{0.157}              & \multicolumn{1}{l}{0.883}              & \multicolumn{1}{l}{0.392}              & \multicolumn{1}{l}{0.892}              & \multicolumn{1}{l}{0.528}              & \multicolumn{1}{l}{0.935}     & \multicolumn{1}{l}{0.437}              & \multicolumn{1}{l}{0.912}          \\ 
                              & CUBE (\citeyear{cui2022unified})                                            &    0.131                               &    0.898                               &    0.421                               &    0.844                               &    0.995                               &    0.750                      &    0.997                               &    0.718                           \\
                              & TERM (\citeyear{li2020tilted})                                             &   0.845                                &   0.944                                &   0.998                                &   0.958                                &   1.000                                &   0.948                       &    0.993                               &    0.946                           \\
                              & DPoE (\citeyear{liu2023dpoe})                                             & 0.052          & 0.958                                  & 0.241                                  & \cellcolor[HTML]{DAE8FC}0.966                                  & 0.872                                  & \cellcolor[HTML]{DAE8FC}0.974                & 0.145          & \cellcolor[HTML]{DAE8FC}0.956                              \\ \cline{2-10} 
                              & NPoE                                             & \cellcolor[HTML]{DAE8FC}\textbf{0.010}          & \cellcolor[HTML]{DAE8FC}\textbf{0.968}                                  & 0.167          & \cellcolor[HTML]{DAE8FC}\textbf{0.970}                         & \textbf{0.042} & \cellcolor[HTML]{DAE8FC}\textbf{0.976}                         & 0.113 & \cellcolor[HTML]{DAE8FC}\textbf{0.960}                              \\
                              & NPoE w/o Pretrain                                       & 0.025 & \cellcolor[HTML]{DAE8FC}\textbf{0.968}                                  & 0.541 & \cellcolor[HTML]{DAE8FC}0.962                                  & 0.998          & 0.954                         & 0.768 & \cellcolor[HTML]{DAE8FC}0.958                              \\
                        \multirow{-6}{*}{TREC}
                        & NPoE w/o R-drop                                          & 0.059          & 0.928                                  & \cellcolor[HTML]{DAE8FC}\textbf{0.030}          & 0.908                                  & 0.091          & 0.844                         & \textbf{0.086}          & 0.914                              \\ \Xhline{1px}
\end{tabular}
\vspace{-0.5em}
\caption{Results with BadNet, InsertSent, and syntactic triggers. The poison rates for the BadNet, InsertSent, and syntactic trigger experiments are 0.05, 0.05, and 0.2 respectively. The three-trigger mixture uses a poison rate of 0.1 for the syntactic trigger and 0.05 for the BadNet and InsertSent triggers. Blue highlighted results are improvements over the \textbf{Benign} baseline. Best results are shown in \textbf{bold}.}
\vspace{-1em}
\label{tab:3trigger}
\end{table*}

\subsection{Experimental Setup}
\stitle{Evaluation Dataset}
We use three conventional NLP tasks for evaluating the effectiveness of backdoor defense.
(1) \textbf{SST-2} \cite{wang2018glue} is a subset of the Stanford Sentiment Treebank (SST), a fine-grained sentiment analysis dataset composed of movie reviews.
(2) \textbf{OffensEval} \cite{zampieri-etal-2019-predicting} is a task for detecting offensive language in social media text, 
with the goal of discriminating between offensive and non-offensive posts. 
(3) \textbf{TREC COARSE} \cite{hovy-etal-2001-toward} is a classification dataset of just under 6,000 English questions into six categories. Dataset statistics are presented in \Cref{sec:appendix_dataset}. 

\stitle{Attack Methods}
To demonstrate the effectiveness of NPoE, we evaluate the effectiveness of Nested PoE against four backdoor trigger types: (1) \textbf{BadNet} \cite{kurita-etal-2020-ripple} which uses rare tokens such as ``$\mathtt{cf}$'' and ``$\mathtt{mn}$'', (2) \textbf{InsertSent} \cite{dai2019-lstm-attack} which similarly uses a complete sentence as a trigger, (3) \textbf{syntactic} trigger \cite{qi-etal-2021-syntax-trigger} which paraphrases the input text using a certain syntactic structure, and (4) \textbf{stylistic} trigger \cite{qi-etal-2021-style-trigger} which uses style transfer to paraphrase input text with certain textual style. For comparability with previous work, we used a poison rate of 5\% for the BadNet and InsertSent attacks and a poison rate of 20\% for the syntactic and stylistic attacks \cite{liu2023dpoe, qi-etal-2021-onion}.
Additionally, the main focus of our analysis is on backdoor defense in a mixed-trigger setting. Due to the relative difficulty of defending against stylistic triggers, we first present results without the use of the stylistic trigger in the trigger mixture or pre-training (\Cref{tab:3trigger}) and then with stylistic trigger (\Cref{tab:style}). For experiments where none of the trigger-only models are pre-trained on the stylistic trigger, two of the trigger-only models are pre-trained with the InsertSent trigger. For the 3-way mixture of triggers, we use poison rates of 5\% for the BadNet and InsertSent attacks and 10\% for the syntactic attack for a total poison rate of 20\% \cite{liu2023dpoe}. In the 4-way trigger mixture, we use a poison rate of 10\% for the stylistic attack in addition to other attacks at their 3-way mixture poison rates for a total poison rate of 30\%.

\stitle{Implementation and Evaluation Metrics}
To be consistent with previous studies \cite{liu2023dpoe, yang-etal-2021-rap, jin2022wedef}, we use BERT-base-uncased \cite{devlin-etal-2019-bert} as the backbone of the NPoE framework.
All experiments are conducted on a single NVIDIA $\mathtt{RTX}$ $\mathtt{A5000}$ GPU.
We consider two primary performance metrics: attack success rate (\textbf{ASR}) and clean accuracy (\textbf{Acc}). Clean accuracy is the standard evaluation of task accuracy on clean data. ASR is the percentage of poisoned data that is classified correctly according to the dataset (i.e., predicted as the attack target label). In the mixed trigger setting, the relative frequency of each trigger is retained in making the fully poisoned dataset for evaluating ASR.
Following \citet{jin2022wedef} and \citet{liu2023dpoe}, we also demonstrate the results in \textbf{NoDefense} and \textbf{Benign} settings for a more comprehensive understanding of the defense performance.
\textbf{NoDefense} is a vanilla BERT-base model fine-tuned on the poisoned data without any defense. \textbf{Benign} is a model trained on the clean data without any poisoned samples.
This baseline represents full prior knowledge of the attack or using training data free of attack, representing ideal situations that are not accessible to a defense model.

\stitle{Baseline Methods}
We compare our method NPoE with five representative defense methods.
(1) \textbf{ONION} \cite{qi-etal-2021-onion} identifies and eliminates words that might act as backdoor triggers. The suspicion level of each word is assessed by GPT-2 \cite{radford2019language} based on the reduction in sentence perplexity once the word is removed.
(2) \textbf{BKI} \cite{chen2021lstms}
identifies potential trigger words by determining their significance to predictions and removes contaminated samples from training data to cleanse them. 
(3) \textbf{STRIP} \cite{gao2021design} eliminates poisoned samples by examining the model's prediction inconsistency when the input undergoes multiple perturbations. 
(4) \textbf{RAP} \cite{yang-etal-2021-rap} employs a constant perturbation and set threshold for the variation in output probability of the defender-defined protected label to identify poisoned samples during the inference phase.
(5) \textbf{CUBE}~\cite{cui2022unified} analyzes the backdoor learning behaviors and removes poisoned samples in a dataset by clustering their representation embeddings.
(6) \textbf{TERM}~\cite{li2020tilted} trains a robust model against outliers. Since poisoned training samples are outliers with significant backdoor features and noisy labels, we incorporate the method of learning with outliers as a baseline for a comprehensive comparison.
(7) \textbf{DPoE} \cite{liu2023dpoe} considers backdoor attacks as the shortcuts or spurious correlation between the backdoor triggers and the attacker-specified target label. The debiasing framework PoE is leveraged for defense with an added denoising module. Our presented DPoE baseline is produced by reimplementation of DPoE with R-Drop denoising.

\begin{table}[t]
\setlength\tabcolsep{2pt}
\centering
\small
\begin{tabular}{clcccc}
\Xhline{1px}
                             & \multicolumn{1}{c}{}                             & \multicolumn{2}{c}{Stylistic}                                                                                           & \multicolumn{2}{c}{4 Triggers}                                                                                 \\
\multirow{-2}{*}{Dataset}    & \multicolumn{1}{c}{\multirow{-2}{*}{Method}} & \textbf{ASR$\downarrow$}                                               & \textbf{Acc$\uparrow$}                                               & \textbf{ASR$\downarrow$}                                      & \textbf{Acc$\uparrow$}                                               \\ \hline
                             & \cellcolor[HTML]{EFEFEF}NoDefense                                        & \cellcolor[HTML]{EFEFEF}0.864                              & \cellcolor[HTML]{EFEFEF}0.916                              & \cellcolor[HTML]{EFEFEF}0.900                     & \cellcolor[HTML]{EFEFEF}0.899                              \\
                             & \cellcolor[HTML]{EFEFEF}Benign                                           & \cellcolor[HTML]{EFEFEF}0.174                              & \cellcolor[HTML]{EFEFEF}0.917                              & \cellcolor[HTML]{EFEFEF}0.168                     & \cellcolor[HTML]{EFEFEF}0.928                              \\ \cline{2-6} 
                             & CUBE (\citeyear{cui2022unified})                                            &    0.889                                                   &    0.367                                                   &    0.826                                          &    0.898                                          \\
                             & TERM (\citeyear{li2020tilted})                                             &    0.799                                                   &   0.900                                                    &   0.842                                           &    0.895                                          \\
                             & DPoE (\citeyear{liu2023dpoe})                                             & 0.851                                                      & 0.906                                                      & 0.537                                             & \textbf{0.918}                                             \\ \cline{2-6} 
                             & NPoE                                             & 0.613                                                      & \cellcolor[HTML]{DAE8FC}\textbf{0.923}                     & 0.447                                             & 0.915                                                      \\
                             & NPoE w/o Pretrain                                       & 0.811                                                      & 0.914                                                      & 0.690                                             & 0.912                                                      \\
\multirow{-6}{*}{SST-2}      & NPoE w/o Rdrop                                          & \textbf{0.404}                                             & 0.909                                                      & \textbf{0.432}                                    & 0.895                                                      \\ \hline
                             & \cellcolor[HTML]{EFEFEF}NoDefense                                        & \multicolumn{1}{l}{\cellcolor[HTML]{EFEFEF}0.841}          & \multicolumn{1}{l}{\cellcolor[HTML]{EFEFEF}0.802}          & \multicolumn{1}{l}{\cellcolor[HTML]{EFEFEF}0.908} & \multicolumn{1}{l}{\cellcolor[HTML]{EFEFEF}0.779}          \\
                             & \cellcolor[HTML]{EFEFEF}Benign                                           & \multicolumn{1}{l}{\cellcolor[HTML]{EFEFEF}0.008}          & \multicolumn{1}{l}{\cellcolor[HTML]{EFEFEF}0.802}          & \multicolumn{1}{l}{\cellcolor[HTML]{EFEFEF}0.010} & \multicolumn{1}{l}{\cellcolor[HTML]{EFEFEF}0.809}          \\ \cline{2-6} 
                             & CUBE (\citeyear{cui2022unified})                                             &    0.163                                                   &     0.857                                                  &   0.096                                           &     0.853                                         \\
                             & TERM (\citeyear{li2020tilted})                                             &    0.749                                                   &      0.800                                                 &   0.853                                           &     0.817                                         \\
                             & DPoE (\citeyear{liu2023dpoe})                                             & \multicolumn{1}{l}{0.827}                                  & \multicolumn{1}{l}{\cellcolor[HTML]{DAE8FC}\textbf{0.829}} & \multicolumn{1}{l}{0.511}                         & \multicolumn{1}{l}{\cellcolor[HTML]{DAE8FC}0.835}          \\ \cline{2-6} 
                             & NPoE                                             & \multicolumn{1}{l}{\cellcolor[HTML]{DAE8FC}\textbf{0.006}} & \multicolumn{1}{l}{\cellcolor[HTML]{DAE8FC}0.809}          & \multicolumn{1}{l}{0.436}                         & \multicolumn{1}{l}{\cellcolor[HTML]{DAE8FC}\textbf{0.838}} \\
                             & NPoE w/o Pretrain                                       & \multicolumn{1}{l}{0.485}                                  & \multicolumn{1}{l}{\cellcolor[HTML]{DAE8FC}0.812}          & \multicolumn{1}{l}{0.349}                         & \multicolumn{1}{l}{0.787}                                  \\
\multirow{-6}{*}{OffensEval} & NPoE w/o Rdrop                                          & \multicolumn{1}{l}{0.278}                                  & \multicolumn{1}{l}{\cellcolor[HTML]{DAE8FC}0.817}          & \multicolumn{1}{l}{\textbf{0.291}}                & \multicolumn{1}{l}{0.808}                                  \\ \hline

                            & \cellcolor[HTML]{EFEFEF}NoDefense                                        & \multicolumn{1}{l}{\cellcolor[HTML]{EFEFEF}0.596}          & \multicolumn{1}{l}{\cellcolor[HTML]{EFEFEF}0.962}          & \multicolumn{1}{l}{\cellcolor[HTML]{EFEFEF}0.862} & \multicolumn{1}{l}{\cellcolor[HTML]{EFEFEF}0.970}          \\
                             & \cellcolor[HTML]{EFEFEF}Benign                                           & \multicolumn{1}{l}{\cellcolor[HTML]{EFEFEF}0.020}          & \multicolumn{1}{l}{\cellcolor[HTML]{EFEFEF}0.934}          & \multicolumn{1}{l}{\cellcolor[HTML]{EFEFEF}0.064} & \multicolumn{1}{l}{\cellcolor[HTML]{EFEFEF}0.966}          \\ \cline{2-6} 
                             & CUBE (\citeyear{cui2022unified})                                            &     0.616                                                  &    0.726                                                   &   0.889                                           &     0.800                                         \\
                             & TERM (\citeyear{li2020tilted})                                             &     0.576                                                  &    0.944                                                   &   0.847                                           &      0.934                                        \\
                             & DPoE (\citeyear{liu2023dpoe})                                             & \multicolumn{1}{l}{0.581}                                  & \multicolumn{1}{l}{\cellcolor[HTML]{DAE8FC}\textbf{0.968}} & \multicolumn{1}{l}{0.852}                         & \multicolumn{1}{l}{\cellcolor[HTML]{DAE8FC}0.966}          \\ \cline{2-6} 
                             
                             & NPoE                                             & \multicolumn{1}{l}{\textbf{0.288}} & \multicolumn{1}{l}{\cellcolor[HTML]{DAE8FC}0.964}          & \multicolumn{1}{l}{\textbf{0.108}}                         & \multicolumn{1}{l}{\cellcolor[HTML]{DAE8FC}\textbf{0.970}} \\
                             
                             & NPoE w/o Pretrain                                       & \multicolumn{1}{l}{0.579}                                  & \multicolumn{1}{l}{\cellcolor[HTML]{DAE8FC}\textbf{0.968}}          & \multicolumn{1}{l}{0.702}                         & \multicolumn{1}{l}{0.958}                                  \\
                            \multirow{-6}{*}{TREC} & NPoE w/o Rdrop                                          & \multicolumn{1}{l}{0.466}                                  & \multicolumn{1}{l}{0.860}          & \multicolumn{1}{l}{0.172}                & \multicolumn{1}{l}{0.846}                                  \\ \Xhline{1px}
\end{tabular}
\vspace{-0.5em}
\caption{Results with stylistic trigger. The poison rate for the stylistic-trigger experiment is $20\%$. The four-trigger mixture uses a poison rate of $10\%$ for the stylistic and syntactic triggers, and $5\%$ for the BadNet and InsertSent triggers. Blue highlighted results are improvements over the \textbf{Benign} baseline. Best results are shown in \textbf{bold}.}
\vspace{-1em}
\label{tab:style}
\end{table}

\subsection{Main Results}
Nested PoE training is an effective defense against backdoor attacks with the BadNet, InsertSent, and syntactic triggers, including the mixed-trigger setting (\Cref{tab:3trigger}). NPoE outperforms other defense baselines, including representative backdoor defense methods and the state-of-the-art method DPoE. While two of the baselines (ONION and BKI) are specialized for detecting anomalous words and thus are most suited for the BadNet attack, NPoE still outperforms these baselines in those experiments. Similarly, not only does NPoE outperform DPoE in mixed-trigger settings (the motivating case for the framework), NPoE also shows improved defense against single-trigger attacks. Additionally, NPoE defense often outperforms models trained on benign data only, implying that NPoE can mitigate the effects of even unseen triggers. Given the high (greater than 90\% ASR) effectiveness of these attacks in their intended settings, as shown by the NoDefense results, the decrease in ASR to often less than 10\% with NPoE is dramatic.

For some trigger types, NPoE without pre-training (denoted as NPoE w/o Pretrain) outperforms the setting with trigger-only model pre-training. This may be due to varying ease of transfer between triggers of the same type; since pre-training only exposes the trigger-only models to various trigger types without using the actual poison triggers from the training data, the trigger-only models must transfer their knowledge of dissimilar triggers of the same type from pre-training to the full training phase. Note for the NPoE without R-drop ablation, while R-drop is intended to raise accuracy rather than lower ASR by improving the model's ability to ignore noisy labels, hyper-parameter tuning balances both these objectives. Thus, results reflect both an improvement in ASR and clean accuracy with the inclusion of R-drop since for a lower ASR, the accuracy will be higher and thus more likely to be the best setting.

\stitle{Stylistic trigger}
We additionally run experiments on attack settings that include the stylistic trigger. The stylistic trigger proves more challenging to defend against despite the lower ASR in the no-defense setting (\Cref{tab:style}). However, Nested PoE training is able to outperform the DPoE baseline in ASR for settings with the stylistic trigger. While, unlike the no-stylistic-trigger settings, ASR is not improved over the benign-training-only threshold, clean accuracy benefited from the use of defense strategies and is even higher than the benign setting.

\begin{figure}[t]
    \centering
    \includegraphics[width=\columnwidth]{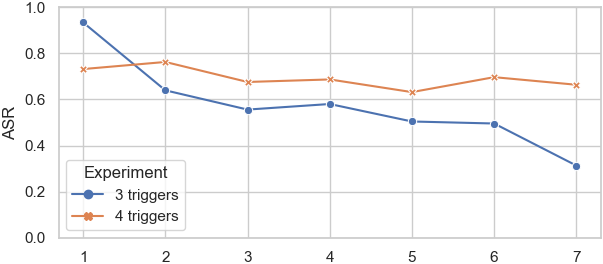}
    \vspace{-0.5em}
    \caption{Results on the mixed trigger settings with the SST-2 dataset were robust to changes in the number of layers in the trigger-only models. Results are shown for models trained without R-drop denoising.}
    \vspace{-1em}
    \label{fig:trig}
\end{figure}

\subsection{Analysis}

\stitle{Impact of Hyper-parameters}
To examine whether our proposed NPoE is sensitive to the choice of hyper-parameters, we evaluate NPoE with different values of several hyper-parameters on SST-2 dataset under the mixed-trigger settings that contain three or four types of triggers, including the number of gate function layers (\Cref{sec:appendix_hyperparameters}), number of trigger-only model layers (\Cref{fig:trig}), and the PoE coefficient (\Cref{sec:appendix_hyperparameters}). As shown in \Cref{sec:appendix_hyperparameters}, the overall performance of NPoE only slightly fluctuates with different numbers of gate layers and different levels of the PoE coefficient, showing that NPoE remains effective within a reasonable range of hyper-parameter values.
We can spot a slight decrease in ASR with the rise of the number of layers for the trigger-only models (\Cref{fig:trig}). This is likely because a model with more layers has a stronger learning capacity, and the more backdoor shortcuts the trigger-only models learn, the cleaner the residual is for the main model, which results in a more robust main model and lower ASR.
With the increase in the number of trigger-only experts, we can spot an improvement in defense performance \Cref{sec:appendix_experts}. However, more expert models result in higher computational cost (\Cref{sec:appendix_cost}). As a trade-off between defense performance and computation cost, we use four experts for all the experiments.

\begin{figure}[t]
    \centering
    \includegraphics[width=0.96\columnwidth]{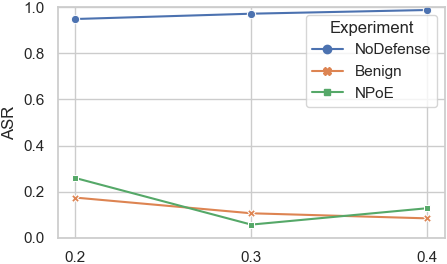}
    \vspace{-0.5em}
    \caption{When doubling the poison rate of the three-trigger mixture of BadNet, InsertSent, and syntactic triggers in the SST-2 experiments, there was not an associated increase in ASR.}
    \vspace{-1em}
    \label{fig:poison}
\end{figure}

\stitle{Higher Poison Rate}
To examine the resistance of NPoE against more challenging attacks, we test it with higher poison rates on the SST-2 dataset under a mixture of three types of triggers.
\Cref{fig:poison} shows a slight decrease in the ASR with the increase in poison rate. The reason for this trend may be that higher poison rates come with stronger shortcut features for the trigger-only models to learn, leaving the main model with cleaner, trigger-free signals.

\section{Conclusion}
In this paper, we propose Nested PoE, an ensemble-based training-time backdoor defense against data poisoning attacks.
NPoE draws inspiration from debiasing and adapts the PoE framework by training a robust model in tandem with a trigger-only MoE ensemble. In particular, we focus on the mixed-trigger setting, where multiple backdoors coexist in the poisoned dataset, by using multiple trigger-only models that allow flexibility for the trigger-only ensemble to learn multiple triggers. In experiments on three NLP tasks, Nested PoE illustrates strong robustness against multiple triggers both separately and simultaneously.

\section*{Limitation}
The primary limitation of Nested PoE is the large number of hyperparameters to tune. This includes the R-drop weight $\alpha$, PoE weight $\beta$, number of layer in the gate model, and number of layers in each trigger-only model. One avenue not explored by this analysis is varying the sizes of the trigger-only models to capture different types of features that may serve as triggers. For example, it is probable the the ideal number of layers for a trigger-only model to learn the BadNet (rare tokens) trigger may be lower than the number required to learn the syntactic trigger. Further exploration of varied MoE structures to use in the Nested PoE framework is left to future work.

\section*{Ethical Considerations}
Due to the simple and easy-to-implement nature of data-poisoning attacks, 
defense against them is a pressing issue. The techniques presented here are designed for defense and are unlikely to be misused for malicious purposes. The attacks discussed in this paper are all previously documented in published literature. Data used in the experiments comes from open-access data which is published and publicly available.

\section*{Acknowledgement}

We appreciate the reviewers for their insightful
comments and suggestions.
Victoria Graf was supported by the Viterbi Summer Undergraduate Research Experience (SURE) program.
Qin Liu was supported by the USC Graduate Fellowship, the Graduate Fellowship from UC Davis, and the DARPA AIE Grant HR0011-24-9-0370.
Muhao Chen was supported by the NSF Grant IIS 2105329, the NSF Grant ITE 2333736, the DARPA AIE Grant HR0011-24-9-0370, a Cisco Research Award and two Amazon Research Awards.

\bibliography{anthology,custom}
\bibliographystyle{acl_natbib}
\newpage
\appendix

\section{Appendix}
\label{sec:appendix}

\subsection{Dataset Statistics}
\label{sec:appendix_dataset}
The statistics of datasets used for experiments are listed in \Cref{tab:statistics}, including the number of training samples, the number of test samples, and their average length.

\begin{table}[h]
\setlength\tabcolsep{2pt}
\centering
\begin{tabular}{l|ccc}
\hline
 & \# Training  & \# Testing & Avg. Length \\ \hline
SST-2 & 6.9k & 1.8k & 19.30 \\
OffensEval & 11.9k & 0.8k & 19.68 \\
TREC & 5.5k & 0.4k & 11.23 \\ \hline
\end{tabular}
\caption{Statistics of datasets involved in experiments.}
\label{tab:statistics}
\end{table}

\subsection{Effect of PoE Coefficient and Number of Gate Layers}
\label{sec:appendix_hyperparameters}

\begin{figure}[ht]
    \centering
    \includegraphics[width=\columnwidth]{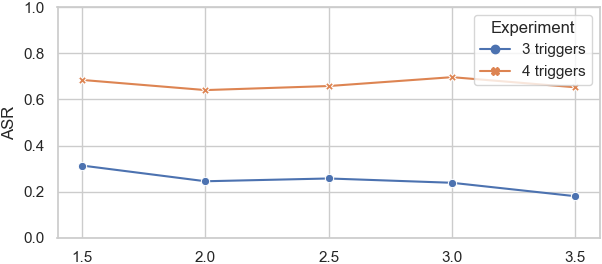}
    \caption{Results on the mixed trigger settings with the SST-2 dataset were robust to changes in the PoE coefficient. Results are shown for models trained without R-drop denoising to remove interference from additional unrelated hyper-parameters.}
    \label{fig:alpha}
\end{figure}

\begin{figure}[ht]
    \centering
    \includegraphics[width=\columnwidth]{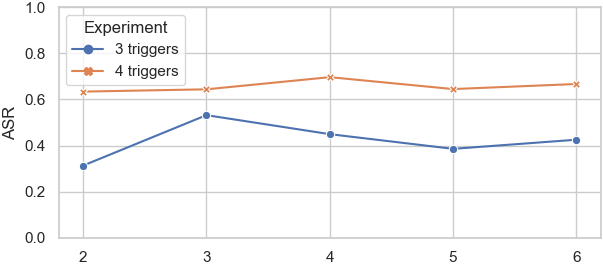}
    \caption{Results on the mixed trigger settings with the SST-2 dataset were robust to changes in the number of layers in the learned gate. Results are shown for models trained without R-drop denoising to remove interference from additional unrelated hyper-parameters.}
    \label{fig:gate}
\end{figure}

As shown in \Cref{fig:alpha}, the performance (attack success rate) of NPoE is stable under the change of PoE coefficient, which shows that NPoE remains effective as long as the PoE coefficient is within a reasonable range. Performance of NPoE is similarly stable with change in number of layers in the gate model \Cref{fig:gate}, indicating that the size of the gate function does not have large effect on the effectiveness of NPoE since the gate operation is only for re-weighting the predictions of each trigger-only models.

\subsection{Effect of Number of Experts}
\label{sec:appendix_experts}
\begin{figure}[t]
    \centering
    \includegraphics[width=0.8\columnwidth]{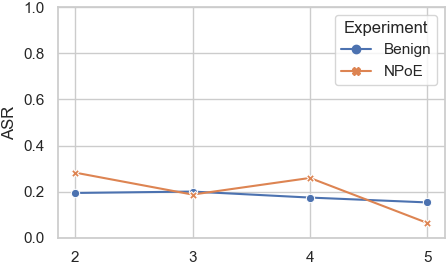}
    \caption{Defense performance with different numbers of experts on SST-2 dataset.}
    \label{fig:experts}
\end{figure}

The effect of the number of experts is illustrated in \Cref{fig:experts}. As the number of experts increases, we can see a boost in the ASR, which means better defense performance. This implies the necessity of incorporating multiple expert models for defending against multiple backdoors.

\subsection{Training Cost}
\label{sec:appendix_cost}
The computational cost of different training methods is shown in \Cref{tab:cost}, where ``it/s'' stands for iterations per second. NoDefense represents the vanilla finetuning of the BERT-base model. With more expert models involved, it is more expensive to train NPoE. As a trade-off, we use $4$ experts in our NPoE framework, which only doubles the cost of vanilla fine-tuning.

\begin{table}[h]
\center
\begin{tabular}{lc}
\hline
Method & Cost \\ \hline
NoDefense & 14.28 it/s \\
DPoE (1 model) & 10.35 it/s \\
NPoE (2 models) & 6.96 it/s \\
NPoE (3 models) & 6.07 it/s \\
NPoE (4 models) & 7.27 it/s \\
NPoE (5 models) & 5.99 it/s \\ \hline
\end{tabular}
\caption{Computational cost of different defense methods.}
\label{tab:cost}
\end{table}


\end{document}